\documentclass{article}
\usepackage{ijcai23}
\usepackage[usenames, dvipsnames]{xcolor}
\usepackage{times}
\usepackage{soul}
\usepackage{url}
\usepackage[hidelinks]{hyperref}
\usepackage[utf8]{inputenc}
\usepackage{caption}
\usepackage{graphicx}
\usepackage{amsmath}
\usepackage{amsthm}
\usepackage{booktabs}
\usepackage{algorithm}
\usepackage{algorithmic}
\usepackage{pdfpages}
\usepackage[switch]{lineno}
\usepackage{soul}
\usepackage{tikz}
\usepackage{fontawesome5}
\usepackage{pgfplots} 
\usetikzlibrary{pgfplots.dateplot}
\usepackage{pgfplotstable}
\usepackage{filecontents}
\usepackage{balance}
\usepackage{booktabs}
\usepackage{adjustbox}

\usepackage{amssymb}
\usepgfplotslibrary{fillbetween}
\usetikzlibrary{patterns}
\usetikzlibrary{backgrounds}
\usetikzlibrary{arrows,automata}
\graphicspath{ {images/} }
\pgfplotsset{compat=newest}
\usepackage{fancyhdr}

\usepackage{hyperref}

\hypersetup{
     colorlinks   = true,
     citecolor    = MidnightBlue,
     linkcolor    = Maroon,
     urlcolor     = Maroon
}

\usepackage[all]{nowidow}
\usepackage{booktabs}
\theoremstyle{[preposition}

\newcommand{\BibTeX}{\rm B\kern-.05em{\sc i\kern-.025em b}\kern-.08em\TeX}

\title{`Why didn't you allocate this task to them?'\\Negotiation-Aware Explicable Task Allocation and Contrastive Explanation Generation}

\author{
Zahra Zahedi$^1$
\and
Sailik Sengupta$^2$\thanks{Work done while at Arizona State University.}\And
Subbarao Kambhampati$^1$
\affiliations
$^1$SCAI, Arizona State University\\
$^2$\faAmazon WS AI Labs\\
\emails
zzahedi@asu.edu,
sailiks@amazon.com,
rao@asu.edu
}

\begin{document}

\maketitle

\begin{abstract}
    Task allocation is an important problem in multi-agent systems. It becomes more challenging when the team-members are humans with imperfect knowledge about their teammates' costs and the overall performance metric. 
    In this paper, we propose a centralized Artificial Intelligence Task Allocation (AITA) that simulates a negotiation and produces a negotiation-aware explicable task allocation. If a team-member is unhappy with the proposed allocation, we allow them to question the proposed allocation using a counterfactual. By using parts of the simulated negotiation, we are able to provide contrastive explanations that provide minimum information about other's cost to refute their foil. With human studies, we show that (1) the allocation proposed using our method appears fair to the majority, and (2) when a counterfactual is raised, explanations generated are easy to comprehend and convincing. Finally, we empirically study the effect of different kinds of incompleteness on the explanation-length and find that underestimation of a teammate's costs often increases it.
\end{abstract}

\vspace{-7pt}
\section{Introduction}

Whether it be assigning teachers to classes \cite{kraus2019ai}, or employees (nurses) to tasks (wards/shifts) \cite{warner1972mathematical}, task allocation is essential for the smooth function of human-human teams. In the context of indivisible tasks, the goal of task allocation is to assign individual agents of a team to a subset of tasks such that a pre-defined set of metrics are optimized. 
When the cost-information about all the team members and a performance measure is known upfront, one can capture the trade-off between some notion of social welfare (such as fairness, envy-free, etc.) and team efficiency (or common rewards) \cite{bertsimas2012efficiency}. In a distributed setting, agents may have to negotiate back-and-forth to arrive at a final allocation \cite{saha2007efficient}. In the negotiation, agents can either choose to accept an allocation proposed by other agents or reject it; upon rejection, agents can propose an alternative allocation that is more profitable for them and, given their knowledge, still acceptable to others. While distributed negotiation-based allocations will at least keep the agents happy with their lot (since they got what they negotiated for), it tends to have two major drawbacks for human negotiators. First, an agent may not be fully aware of their teammates' costs and the performance metrics, resulting in the need for iteratively sharing cost information. Second, the process can be time-consuming and can increase the human's cognitive overload, leading to sub-optimal solutions.\\ \looseness = -1
In contrast, a centralized allocation can be efficient, but will certainly be contested by disgruntled agents who, given their incomplete knowledge, may not consider it to be acceptable. Thus, a centralized system needs to be ready to provide the user with an explanation. As discussed in \cite{kraus2019ai}, such explanations can aid in increasing people's satisfaction \cite{bradley2009dealing} and maintain the system's acceptability \cite{miller2018explanation}. In a multi-agent environment such as task allocation, providing explanations is considered to be both important and a challenging problem \cite{kraus2019ai}.
\looseness = -1
\begin{figure}[!ht]
    \centering
    \includegraphics[width=.9\columnwidth]{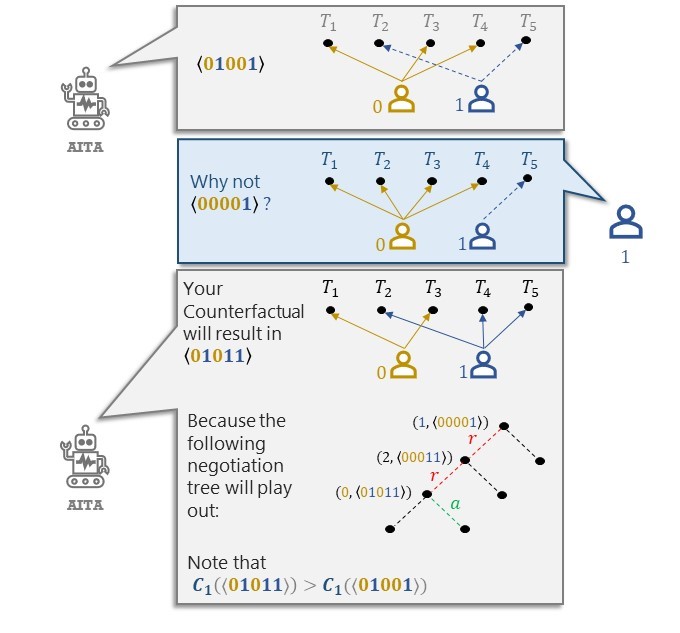}
    \caption{AI Task Allocator (AITA) comes up with a negotiation-aware explicable allocation $\langle01001\rangle$ for a set of two humans-- $0$ and $1$. In this allocation, human $0$ is assigned tasks $1, 3$ and $4$ and agent $1$ is assigned tasks $2$ and $5$. A dissatisfied human $1$ questions AITA with a counterfactual allocation $\langle00001\rangle$, where he/she just needs to do task $5$ (they believe task $5$ is much more difficult and will take similar effort compared to doing all the 4 others). AITA then explains why the original proposed allocation (i.e. $\langle01001\rangle$) is better than the counterfactual allocation (i.e. $\langle00001\rangle$). The graph of the negotiation tree can be given as a dialogue "if human $1$ proposes the allocation $\langle01001\rangle$, it will be rejected and AITA will offer $\langle00011\rangle$, which will then be rejected and human $0$ will propose a counter offer $\langle01011\rangle$ which will then will have to be accepted by all. This final allocation would have a higher cost for you (human $1$) than the first proposed allocation. Hence, the counterfactual allocation will eventually result in worse-off allocation for human $1$.}
    \label{fig:aim_concept}
    \vspace{-.5cm}
\end{figure}

To address these challenges, we blend aspects of both the (centralized and distributed) approaches and propose AITA, an Artificial Intelligence-powered Task Allocator. Our system (1) uses a centralized allocation algorithm patterned after negotiation to come up with an allocation that explicitly accounts for the costs of the individual agents and overall performance, and (2) can provide contrastive explanation when a proposed allocation is contested using a counterfactual. We assume AITA is aware of all the individual costs and the overall performance costs.\footnote{The proposed methods work even when this assumption is relaxed and AITA's proposed allocation and explanation initiate a conversation to elicit human preferences that were not specified upfront. We plan to consider this longitudinal aspect in the future.} Use of a negotiation-based mechanism for coming up with a negotiation-aware explicable allocation helps reuse the inference process to provide contrastive explanations. Our explanations have two desirable properties. First, the negotiation-tree based explanation by AITA has a graphical form that effectively distills relevant pieces from a large amount of information (see \autoref{fig:aim_concept}); this is seen as a convenient way to explain information in multi-agent environments \cite{kraus2019ai}.
Second, the explanation, given it is closely tied to the inference process, acts as a certificate that guarantees explicability to the human (i.e. no other allocation could have been more profitable for them while being acceptable to others).
While works like \cite{kraus2019ai} recognize the need for such negotiation-aware and contestable allocation systems, their work is mostly aspirational. To the best of our knowledge, we are the first to provide concrete algorithms for the generation and explanation of allocation decisions.
To evaluate our work, we conduct human studies in three different task allocation scenarios and show that the allocations proposed by AITA are deemed fair by the majority of subjects. Users who questioned AITA's allocations, upon being explained, found it understandable and convincing in two control cases. Further, we consider an approximate version of the negotiation-based algorithm for larger task allocation domains and, via numerical simulation, show how underestimation of a teammate's costs and different aspects of incompleteness affect explanation length.
\vspace{-7pt}
\section{Related Works}
In this section, we situate our work in the landscape of multi-agent task allocation, explicable decision making and model reconciliation explanations.

\noindent\textbf{Task allocation\quad} 

As mentioned above, challenges in task allocation are either solved using centralized or distributed approaches; the later are more considerate of incomplete information settings. Centralized approaches often model the allocation problem as a combinatorial auction
\cite{hunsberger2000combinatorial,cramton2006introduction}. In human teams, members often have incomplete knowledge about fellow team members and overall performance costs. Thus, a proposed allocation may not seem reasonable to the human. Given existing centralized approaches, there is no way for the human to initiate dialog. Moreover, regardless of optimality or accuracy, a centralized decision making system should be contestable
On the other hand, distributed methods allow agents to autonomously negotiate and agree on local deals \cite{chevaleyre2010simple,chevaleyre2006issues}, finally to reach a consensus-- an allocation that is Pareto Optimal \cite{brams1996fair
,saha2007efficient,endriss2006negotiating,endriss2003optimal}.
Beyond negotiation mechanisms, works have also explored the use of bargaining games to model bilateral (i.e. two-agent) negotiation scenarios \cite{erlich2018negotiation,fatima2014negotiation,peled2013agent}. 

\noindent\textbf{Explicable Decision Making}

Effective human-aware AI systems should generate decisions that are aligned with human expectations. In this regard, explicability is defined based on the alignment of an AI agent's decision to a human's expectations \cite{zhang2017plan}. For planning problems, there exist numerous ways, such as syntactic difference, learned labeling function, and cost difference, to calculate this alignment \cite{kulkarni2019explicable,zhang2017plan,olmo2020not,sreedharan2019expectation}.
While \cite{sengupta2018ma} investigates explanations when planning with a team of humans, none of these works consider explicability when the decision is of interest to multiple humans who have individual preferences and expectations about other humans. In our work, we address this challenge and try to come up with a negotiation-aware explicable allocation that is within a bounded distance of all human's optimal allocation and Pareto optimal in nature.


\noindent\textbf{Model Reconciliation Explanations \quad}

\looseness=-1
Human-aware AI systems are often expected to be able to explain their decisions
\cite{miller2018explanation}.
To generate the various types of explanations, several computation methods exist in AI, ranging from explaining decisions of machine learning systems \cite{ribeiro2016should,melis2018towards,rosenfeld2019explainability,anjomshoae2019explainable,mualla2022quest} to explaining sequential decisions by planners \cite{chakraborti2020emerging}. In \cite{lyons2021conceptualising,van2021evaluating}, the authors identify the need for allowing dialog and being able to provide causal explanations as core principles of explanation in decision making systems. While recent works enable humans to contest black-box decisions \cite{kluttz2022shaping,aler2020contestable} and provide causal explanation for an observed behavior \cite{madumal2020explainable}, our work is the first to exhibit both these characteristics in the context of a task-allocation system with multiple humans stakeholders. 
The need for explanations in our work arises out of the incomplete knowledge the human has about their teammates and the team's performance metric. 
We subscribe to the idea of model reconciliation as explanations and enable AITA (similar in spirit to what \cite{sreedharan2018hierarchical} for planning problems) to come up with contrastive explanation when the human deems the proposed task-allocation inexplicable and contests it with a counterfactual.
\vspace{-7pt}
\section{Problem Formulation}
Task allocation problems are categorized as mixed-motive situations for humans (especially in setting where agents are ought to fulfill their tasks and cannot dismiss it). They are cooperative in forming a team but in getting the assignment they are selfish and considering their own interest. So, for task assignment humans are selfish but at the same time in order to hold a bargain in the team and keep the team they need to consider other teammates and be cooperative. In other words, they are selfish but there is a bound to their selfishness, and the bound comes so the team will not be broken due to selfishness. \\ \looseness =-1
Our task allocation problem can be defined using a $3$-tuple $\langle A, T, C \rangle$ where $A=\{0, 1, \dots, n\}$ where $n$ denotes the AI Task Allocator (AITA) and $0, \dots n-1$ denotes the $n$ humans, $T = \{T_1, \dots, T_m\}$ denotes $m$ indivisible tasks that need to be allocated to the $n$ humans, and $C = \{C_0, C_1, \dots C_n\}$ denotes the cost functions of each agent (preferences over tasks, skills and capability of doing tasks are characterized in cost function-- eg. cost of infinity for being incapable in doing a task). $C_n$ represents the overall performance cost metric associated with a task allocation outcome $o$ (defined below).

For a task $t$, we denote the human $i$'s cost for that task as $C_i(t)$. Let $O$ denote the set of allocations and an allocation $o (\in O)$ represent a one-to-many function from the set of humans to tasks; thus, $|O| = n^{m}$. An outcome $o$ can be written as $\langle o_1, o_2, \dots o_m \rangle$ where each $o_i \in \{0,\dots,n-1\}$ denotes the human performing task $i$. Further, let us denote the set of tasks allocated to a human $i$, given allocation $o$, as $T_i = \{j: o_j = i\}$. For any allocation $o \in O$, there are two types of costs for AITA:

\noindent $(1)$ Cost for each human $i$ to adhere to $o$. In our setting, we consider this cost as $C_i(o) = \Sigma_{j\in T_i} C_i(j)$.

\noindent $(2)$ An overall performance cost $C_n(o)$.

\noindent
\textit{Example.~~} Consider a scenario with two humans $\{0,1\}$ and five tasks $\{t_1, t_2, t_3, t_4, t_5\}$. An allocation outcome can thus be represented as a binary (in general, base-$n$) string of length five (in general, length $m$). For example, $\langle 01001 \rangle$ represents a task allocation in which agent $0$ performs the three tasks $T_0 = \{t_1, t_3, t_4\}$ and $1$ performs the remaining two tasks $T_1 = \{t_2, t_5\}$. The true cost for human $0$ is $C_0(\langle 01001 \rangle) = C_0(t_1) + C_0(t_3) + C_0(t_4)$, while the true cost for $1$ is $C_1(t_2) + C_1(t_5)$.


\vspace{0.03cm}
\noindent
\textbf{Negotiation Tree~~}
A negotiation between agents can be represented as a tree whose nodes represent a two-tuple $(i, o)$ where $i \in A$ is the agent who proposes outcome $o$ as the final-allocation. In each node of the tree, all other agents $j\in A \setminus {i}$ can choose to either {\em accept} or {\em reject} the allocation offer $o$. If any of them choose to reject $o$, in a turn-based fashion\footnote{Note that there is an ordering on agents offering allocation (upon rejection by any of the agents). AITA offers first, followed by each team member in some order and then it continues in a round-robin fashion.}, the next agent $i+1$ makes an offer $o'$ that is
(1) not an offer previously seen in the tree (represented by the set $O_{parents}$, and
(2) is optimal with regards to agent $i+1$'s cost among the remaining offers $O\setminus{O_{parents}}$. This creates the new child  $(i+1, o')$ and the tree progresses either until all agents {\em accept} the offer or all outcomes are exhausted. Note that in the worst case, the negotiation tree can consist of $n^m$ nodes, each corresponding to one allocation in $O$. Each negotiation step, represented as a child in the tree, increases the time needed to reach a final task allocation. Hence, similar to \cite{baliga1995multilateral}, we consider a discount factor (given we talk about costs as opposed to utilities) as we progress down the negotiation tree.\\
Although we defined what happens when an offer is rejected, we did not define the criteria for rejection.
The condition for rejection or acceptance of an allocation $o$ can be defined as follows.%
\begin{eqnarray}
\begin{cases}
accept~ o & if ~ C_i(o) \leq C_i(O_{na\_exp}^i)\\ 
reject~ o & otherwise
\end{cases}
\nonumber
\end{eqnarray}
where $O_{na\_exp}^i$ represents a {\em negotiation-aware explicable allocation} as per agent $i$.
\vspace{-10pt}
\section{Proposing a Negotiation-Aware Explicable Allocation}
In this section, we first formally define a negotiation-aware explicable allocation followed by how it can be computated.

\noindent \textbf{Definition--Negotiation-Aware Explicable Allocation: }
An allocation is considered explicable by all agents iff, upon negotiation, all the agents are willing to accept it. Formally, an acceptable allocation at step $s$ of the negotiation, denoted as $O_{na\_exp}(s)$, has the following properties:\\
    1. All agents believe that allocations at a later step of the negotiation will result in a higher cost for them.
    \[
    \forall i,~\forall s^{'} > s \quad \hspace{5pt}C_i(O(s^{'}))> C_i(O_{na\_exp}(s))
    \]
    2. All allocations offered by agent $i$ at step $s''$ before $s$, denoted as $O^i_{opt}(s'')$, is rejected at least by one other agent. The $opt$ in the subscript indicates that the allocation $O^i_{opt}(s'')$ at step $s''$ has the optimal cost for agent $i$ at step $s''$. Formally,
    \[
    \forall s^{''} < s,~\exists j \not = i,\quad  C_j(O_{opt}^i(s^{''}))>C_j(O_{na\_exp}(s))
    \]
We now describe how AITA finds a negotiation-aware explicable allocation.

\vspace{0.05cm}
\noindent
\textbf{Negotiation-Aware Explicable Allocation Search~~}
The negotiation process to find an explicable allocation can be viewed as an sequential bargaining game. At each period of the bargaining game, an agent offers an allocation in a round-robin fashion. If this allocation is accepted by all agents, each agent incurs a cost corresponding to the tasks they have to accomplish in the allocation proposed (while AITA incurs the team's performance cost). Upon rejection (by even a single agent), the game moves to the next period.
Finding the optimal offer (or allocation in such settings) needs to first enumerate all the periods of the sequential bargaining game. In our case, this implies constructing an allocation enumeration tree, i.e. similar to the negotiation tree but considers what happens if all proposed allocations were rejected. In the generation of this allocation enumeration tree, we assume the humans, owing to limited computational capabilities, can only reason about a subset of the remaining allocations. While any subset selection function can be used in all the algorithms presented in this paper, we will describe a particular one in the next section.\looseness = -1

Given that the sequential bargaining game represents an extensive form game, the concept of Nash Equilibrium allows for non-credible threats. In such settings a more refined concept of Sub-game Perfect Equilibrium is desired \cite{osborne2004introduction}. We first define a sub-game and then, the notion of a Sub-game Perfect Equilibrium.
\\[0.4em]
{\em
\noindent \textbf{Sub-game:} After any non-terminal history, the part of the game that remains to be played (in our context, the allocations not yet proposed) constitutes the sub-game.
}
\\[0.4em]
{\em
\noindent \textbf{Sub-game Perfect Equilibrium (SPE):}
A strategy profile $s^{*}$ is the SPE of a perfect-information extensive form game if for every agent $i$ and every history $h$ (after which $i$ has to take an action), the agent $i$ cannot reduce its cost by choosing a different action, say $a_i$, not in $s^{*}$ while other agents stick to their respective actions. If $o_h(s^{*})$ denotes the outcome of history $h$ when players take actions dictated by $s^{*}$, then $C_i(o_h(s^{*})) \leq C_i(o_h(a_i, s^{*}_{-i})$.
}

Given the allocation enumeration tree, we can use the notion of {\em backward induction} to find the optimal move for all agents in every sub-game \cite{osborne2004introduction}. We first start from the leaf of the tree with a sub-tree of length one. We then keep moving towards the root, keeping in mind the best strategy of each agent (and the allocation it leads to). We claim that if we repeat this procedure until we reach the root, we will find a negotiation-aware explicable allocation. To guarantee this, we prove two results-- (1) an SPE always exists and can be found by our procedure and (2) the SPE returned is an explicable allocation.

{\em
\noindent \textbf{Lemma} There exists a non-empty set of SPE. An element of this set is returned by the backward induction procedure.
}

\noindent \textit{Proof Sketch.}
Note that the backward induction procedure always returns a strategy profile; in the worst case, it corresponds to the last allocation offered in the allocation enumeration tree. Each agent selects the optimal action at each node of the allocation enumeration tree. As each node represents the history of actions taken till that point, any allocation node returned by backward induction (resultant of optimal actions taken by agents), represents a strategy profile that is an SPE by definition. Thus, an SPE always exists. \hfill $\square$

{\em
\noindent \textbf{Corollary} The allocation returned by backward induction procedure is an acceptable allocation.
}

\noindent \textit{Proof Sketch.} A proof by contradiction shows that if the allocation returned is not an acceptable allocation, then it is not the SPE of the negotiation game, contradicting {\em Lemma}. \hfill $\square$

\noindent
Given the introduced algorithm, AITA, with the correct information about the costs, and considering the limited computational capability of the humans, cames up with a negotiation-aware explicable allocation and proposed it to the human.
\vspace{-10pt}
\section{Counterfactual Allocation and Explaining a Proposed Allocation}
\vspace{-3pt}
\noindent
\textbf{Counterfactual Allocation}
In scenarios where the human has (1) access to all the true costs of them and others and (2) computation capabilities to reason about a full negotiation tree, they will simply realize that AITA's allocation is explicable and does not need an explanation.
However, for many real world settings, a human may not be fully aware of the utility functions of the other humans \cite{saha2007efficient}. So in our setting, we formalize this by assuming a human $i$ is only aware of its costs $C_i$ and has noisy information about all the other utility functions $C_j ~\forall~ j\neq i$ and the performance costs (when $j = n$). We represent $i$'s noisy estimate of $j$'s cost as $C_{ij}$. For a task $t$, we denote the human $i$'s perception of $j$'s cost as $C_{ij}(t)$. Given the incomplete information setting, a human's perception of costs for an allocation $o$ relates to their (true) cost $C_i(o)$, noisy idea of other agent's costs $C_{ij}(o)$, and noisy idea of the overall team's performance cost $C_{in}(o)$. Given human $i$'s limited computational abilities, and knowledge about $C_i, C_{ij} (\forall i \neq j)$, and $C_{in}$ (the latter two being inaccurate), $O_{na\_exp}^i$ might not be equal to $O_{na\_exp}$ that AITA proposes. Therefore, a counterfactual allocation may raise by human $i$ who may be able to come up with an allocation that they think has lower cost for them and will be {\em accepted} by all other players. Note that human may assume the centralized agent may make inadvertent errors but not deliberate misallocation and there is no consideration about deception, misuse or irrationality in this work.
Formally, we define the notion of an optimal counterfactual as follows.

\vspace{0.03cm}
{\em
\noindent \textbf{The optimal counterfactual} for a human $i$ is an alternative allocation $o'$, given AITA's proposed allocation $o$, that has the following properties.
\begin{enumerate}
    \item $o'$ is in the set of allocations regarding their limited computational capability \hfill \textcolor{gray}{(this implies that $o \neq o'$)}
    \item $C_i(o) > C_i(o')$ \hfill \textcolor{gray}{($i$ has lower cost in $o'$)}
    \item $o'$ is an SPE in the allocation enumeration tree made from allocations from $o$ given their computational capability.
\end{enumerate}
}


\vspace{0.03cm}
\noindent
\textbf{Explanation as a Negotiation Tree}

We show that when an optimal counterfactual $o_i$ exists for a human $i \in \{0, \dots, n - 1\}$, AITA can come up with an effective explanation that refutes it, i.e. explains how the counterfactual $o_i$ can later result in an acceptable allocation that is not better than $o$.

We thus propose to describe a negotiation tree, which starts with the counterfactual $o_i$ at its root and excludes the original allocation $o$, as an explanation. This differs from the human's negotiation tree because it uses the actual costs as opposed to using the noisy cost estimates.\footnote{Noisy estimates of other agents are not needed to be known by AITA in order to generate explanation.} Further, AITA, with no limits on computation capabilities, can propose allocations that are not bound by the subset selection function.
We finally show that an SPE in this tree results in an SPE that cannot yield a lower cost for the human $i$ than $o$. At that point, we expect a rational human to be convinced that $o$ is a better allocation than the counterfactual $o_i$. Note that a large allocation problem does not imply a long explanation (i.e. a negotiation tree of longer path from root to lowest leaf). In turn, a larger problem does not necessarily make an explanation verbose. We can now define what an explanation is in our case.

{\em
\noindent \textbf{Explanation. }
An explanation is a negotiation tree (note that this negotiation tree can be cast as a natural language or dialogue) with true costs that shows the counterfactual allocation $o_i$ will result in a final allocation $\hat{o}_i$ such that $C_i(\hat{o}_i) \geq C_i(o)$. 
}

Even before describing how this looks in the context of our example, we need to ensure one important aspect-- given a counterfactual $o'$ against $o$, an explanation always exists.

{\em
\noindent \textbf{Proposition}
Given allocation $o$ (the explicable allocation offered by AITA) and a counterfactual allocation $o_i$ (offered by $i$), there will always exist an explanation.
}\\[0.4em]
\noindent \textit{Proof Sketch:}
We showcase a proof by contradiction; suppose an explanation does not exist. It would imply that there exists $\hat{o}_i$ that reduces human $i$'s cost (i.e. $C_i(o) \geq C_i(\hat{o}_i)$) and is accepted by all other players after negotiation. By construction, $o$ was an explicable allocation and thus, if a human was able to reduce its costs without increasing another agent's cost, all agents will not have accepted $o$.
As $\hat{o}_i$ is also the resultant of a sub-game perfect equilibrium strategy of the allocation enumeration tree with true costs, AITA would have chosen $\hat{o}_i$. Given AITA chose $o$, there cannot exist such a $\hat{o}_i$.\hfill $\square$

\vspace{0.03cm}
\noindent
\textbf{An Example of Limited Computational Capabilities: Subset Selection Function}
An example of subset function can be assuming that given a particular allocation outcome $o=\langle o_1, \dots o_j \rangle$, a human will only consider outcomes $o'$ where only one task in $o$ is allocated to a different human $j$. In the context of our example, given allocation $\langle 010 \rangle$, the human can only consider the three allocations $\langle 011 \rangle$, $\langle 000 \rangle$ and $\langle 110 \rangle$; outcomes are one Hamming distance away. With this assumption in place, a human is considered capable of reasoning about a negotiation tree with $m*(n-1)$ allocations (as opposed to $n^m$) in the worst case.\footnote{As specified above, other ways to limit the computational capability of a human can be factored into the backward induction algorithm.}  
While there can be other ways to assume subset function that shows human limited computational capability. The described 1-edit distance function will be used in Experimental Results as an assumption for limited computational capability of human.
\vspace{-10pt}
\section{Experimental Results}
\vspace{-5pt}
In this section, we try to examine whether the proposed explicable allocation is perceived to be fair and the effectiveness of the contrastive explanations generated by AITA. For this, we consider two different human study experiments. In one setting, if a user selects an allocation is unfair, we ask them to rate our explanation and two other baselines (relative case). In the other case, we simply provide them our explanation (absolute case). We then consider the effect of different kinds of inaccuracies (about costs) on the explanation length for a well-known project management domain \cite{certa2009multi}.

\noindent
\textbf{Human Subject Studies} We briefly describe the study setup for the two human studies.

\begin{table}[tp]
\centering
\begin{tabular}{l p{0.9cm} p{1.85cm} p{1.85cm}}
\toprule
Domain &\multicolumn{3}{c}{Percentage of selected options}\\[0.45em]
& Fair & Optimal \newline Counterfactual & Sub-optimal \newline Counterfactual\\
\cmidrule{2-4}
Cooking & $84.2\%$ & $15.8\%$ & $0\%$ \\
Class Project & $86.4\%$ & $7.9\%$ & $5.3\%$ \\
Paper Writing & $55\%$ & $37.5\%$ & $7.5\%$ \\
\bottomrule
\end{tabular}
\caption{Options selected by participants for the three different domains used in the human studies.}
\label{fig:fair_cs_selection}
\vspace{-1em}
\end{table}


\noindent\textit{\textbf{Relative Case~~}} In this setting, we consider two different task allocation scenarios. The first one considers cooking an important meal at a restaurant with three tasks and two teammates-- the chef and the sous-chef. The second one considers dividing five tasks associated with a class project between a senior grad and a junior grad/undergrad student. In the first setting, the human subject plays the role of a sous-chef and are told they are less efficient at cooking meals and may produce a meal of low overall quality (the performance cost is interpreted as the customer ratings seen so far). In the second setting, the human subject fills in the role of the senior student who is more efficient at literature review, writing reports and can yield a better quality project (as per the grading scheme of an instructor). We recruited a total of $38$ participants of whom $54.3\%$ were undergraduate and $45.7\%$ were graduate students in Computer Science \& Engineering and Industrial Engineering at our university
. All of them were made to answer a few filter questions correctly to ensure they fully understood the scenarios.

In the study, we presented the participants with AITA's proposed allocation and counterfactual allocations that adhere to the one-hamming distance subset selection function defined above. This let us consider two and three counterfactual allocations for the cooking and the class project domains respectively.\footnote{A detailed description of the domains and the study can be found in the supplementary material}. When the human selects a counterfactual, implying that AITA's proposed allocation is inexplicable, we present them with three explanations. Besides our negotiation-tree based explanation, we showcase two baseline explanations-- (1) A {\em vacuous} explanation that simply states that the human's counterfactual won't be accepted by others and doesn't ensure a good overall performance metric, (2) A {\em verbose} explanation that provides the cost of all their teammates and the performance metric for all allocations.

\noindent\textit{\textbf{Absolute Case Setup~~}}
In this case, we considered a task allocation scenario where two research students-- a senior researcher and a junior researcher-- are writing a paper and the three tasks relate to working on different parts of the paper. In this setting, we gathered data from $40$ graduate students. Similar to the previous case, the subjects have to select whether the AITA's proposed allocation is fair or select between either of the two counterfactual allocations (each adhering to the subset selection constraint). In contrast to the previous case, upon selecting a counterfactual, the subject is only presented with the negotiation-tree explanation.

\noindent{\textit{\textbf{Results}}~~} In \autoref{fig:fair_cs_selection}, we see that a majority of the participants selected AITA's allocation as fair across all the three different domains. This shows that our formally defined negotiation-aware explicable allocation does indeed appear fair to participants. Given that a set of the participants felt that the allocation is unfair, we were able to establish a scenario where they desired explanations. Instead of having them calculate a counterfactual, we presented them with options they would might want to use as a counterfactual. We noticed that the highest selected counterfactual was the optimal counterfactual, calculated using the SPE mechanism over the human's negotiation tree in the background (that is generated assuming the human's computational capabilities are limited). For the cooking and paper writing domains, the result was statistically significant, highlighting that our computational methods are indeed able to come up with the optimal counterfactual that is in line with how humans would come up with a counterfactual. The least selected option was the other sub-optimal counterfactual allocations.\\ \looseness=-1
For participants who asked for explanations by providing a counterfactual, we had two settings-- relative and absolute. In the relative case, they were asked to rate the comprehensibility and convincing power of the three explanations on a scale of $1-5$. In absolute case, only our explanation was provided to them. We observed that the negotiation tree was judged to be {\em understandable} and {\em moderately convincing} on average in the absolute setting. In the cumulative setting, results in both the cooking and the class project domain show that our explanation is the most convincing one. It is also perceived as the most understandable explanation, but shared the stage with the Vacuous explanation (the average scores and additional metrics are available in the supplementary file).

\noindent{\textit{\textbf{Statistical Significance}}~~}
To further compare our negotiation-tree based explanation with the other two-- vacuous and verbose-- explanations, we performed one-way ANOVA and a non-parametric Kruskal-Wallis tests with Bonferroni correction. The results of both ANOVA test ($F(2,24) = 8.96$, $p = 0.001$), and Kruskal-Wallis test ($p= 0.003$) show that (1) the three explanations are  different in term of convincing power and (2) the negotiation-tree based explanation is the most convincing while the vacuous explanation is the least convincing (rank sum: $172.5 > 140 > 65.5$). However, ANOVA ($F(2,24) = 0.04$, $p = 0.96$) and Kruskal-Wallis ($p = 0.68$) tests show that all the three explanations are not greatly different when it comes to human understanding. So, for understanding, we further performed pair-wise comparison of explanations with pairwise Mann-Whitney tests. Along similar lines, this test also didn't show any significant difference between the pairs ($p = 0.63 $, $0.42$ and $0.73$). Finally, we used the TOST (equivalent test) to evaluate if three explanations are equivalent in terms of understanding. TOST showed that all three explanations are all equal in terms of understanding with $90\%$ confidence interval $(-0.7 \hspace{5pt} 1.2)$ and $(-0.6 \hspace{5pt}0.8)$ in $(-1.5 \hspace{5pt}1.5)$. Therefore, we can conclude that while \textit{our negotiation-tree based explanation is the most convincing one}, it is similar to the other in terms of understandable rating by humans.

\noindent
\textbf{Impact of Noise on Explanations}
For this study, we use the project management scenario from \cite{certa2009multi} in which human resources are allocated to \textit{R\&D} projects. We modify the original setting in three ways.
First, we consider two and four humans instead of eight for assigning the five projects, allowing a total of $2^5 = 32$ possible allocations. It allows for explanations of reasonable length where allocations can be represented as 5-bit binary strings (see \autoref{fig:aim_concept}).
Second, we only consider the skill aspect, ignoring the learning abilities of individuals and social aspects of an allocation. This was mostly done because we could not confidently specify a relative prioritization of the three. We use the skill to measure the time needed, and in turn the cost, for completing a project (more the time needed, more the cost). There are a total of $2*5 = 10$ actual costs, $5$ for each human (the detail costs in supplementary material), and $10$ additional costs representing the noisy perception of one human's cost by their teammate.
Third, we consider an aggregate metric that considers the time taken by the two humans to complete all the tasks. Corresponding to each allocation, there are $32$ (true) costs for team performance. With these costs, as shown in \autoref{fig:aim_concept}, the negotiation-aware explicable allocation is $\langle 01001 \rangle$, the optimal counterfactual for agent $1$ is $\langle 00001 \rangle$ which is revoked by AITA using an explanation tree of length three.

\noindent{\textit{\textbf{Impact of Norm-bounded Noise.}}~~}
The actual cost $C_i$ of each human $i$ as a vector of length $m$. A noisy assumption can be represented by another vector situated $\epsilon$ (in $l_2$ norm) distance away. By controlling $\epsilon$, we can adjust the magnitude of noise a human has. In Figure \ref{fig:eps_el}, we plot the effect of noise on the average explanation length. The noisy cost vectors are sampled from the $l_2$ norm ball within $\epsilon$ radius scaled by highest cost in the actual cost vectors \cite{calafiore1998uniform}.\footnote{We clip the norm-ball to be in the positive quadrant as negative costs are meaningless in our setting.}
The y-axis indicates the length of the replay negotiation tree shown to the human. Even though the maximum length of explanation could be $31(2^5-1)$, we saw the maximum explanation length was $8$.
Given that every noise injected results in a different scenario, we average the explanation length across ten runs (indicated by the solid lines). We also plot the additive variance (indicated by the dotted lines). The high variance on the negative side (not plotted) is a result of the cases where either (or both) of the human(s) human team members didn't have an optimal counterfactual and thus, the explanation length was zero.

\begin{figure}[t]
\begin{tikzpicture} 
\begin{axis}[
    width=.85\linewidth,
	height=.65\linewidth,
    ymajorgrids=true,
    xmajorgrids=true,
    grid style=dashed,
    title={},
    xmin=-0.5,
    ymin=-0.5,
    xlabel={$\epsilon \rightarrow$}, 
    ylabel={Explanation length $\rightarrow$},
    legend style={
        at={(1.15,0.42)},
        font=\small
    }
  ]
\addplot[name path= dr, color=Bittersweet, smooth,mark=o] coordinates{ (0,1) (1,1.409090909) (2,1.636363636) (3, 1.909090909) (4, 1.545454545) (5,1.818181818) (6, 2.045454545) (7, 1.818181818)}; 
\addlegendentry{\tiny$\mu_{rand}$}

\addplot+[name path= upr, color=Bittersweet!70!white, dotted, smooth, mark=x] plot coordinates{ (0,1) (1,1.409090909+1.848897253) (2,1.636363636+1.846761034) (3,1.909090909+1.880649384) (4,1.545454545+1.657518754) (5,1.818181818+1.732050808) (6, 2.045454545+1.899445903) (7, 1.818181818+1.838191331)};
\addlegendentry{\tiny$\mu_{rand} + \sigma_{rand}$}

\addplot[name path= d, color=TealBlue, smooth,mark=square] coordinates{ (0,1) (1,1.4) (2,1.6) (3, 1.65) (4, 1.95) (5,2.1) (6, 2.2) (7, 2.4)}; 

\addlegendentry{\tiny$\mu_{PN}$}

\addplot+[name path= up, color=TealBlue!50!black, dotted, smooth, mark=x] plot coordinates{ (0,1) (1,1.4+1.846761034) (2,1.6+1.848897253) (3, 1.65+1.899445903) (4, 1.95+1.788854382) (5,2.1+1.803505359) (6, 2.2+1.880649384) (7, 2.4+1.667017507)};

\addplot[color=Bittersweet!12!white, fill opacity=0.5] fill between[of=upr and dr];

\addplot[color=TealBlue!20!white, fill opacity=0.5] fill between[of=up and d];

\addlegendentry{\tiny$\mu_{PN} + \sigma_{PN}$}
\end{axis}
\end{tikzpicture}
\vspace{-5pt}
\caption{Mean length of explanations as the amount of noise added to the actual costs increases.}
\label{fig:eps_el}
\vspace{-0.55cm}
\end{figure}

We initially hypothesized, based on intuition, that an increase in the amount of noise will result in a longer explanation. The curve in red (with $\circ$) is indicative that this is not true. To understand why this happens, we classified noise into two types-- Optimistic Noise (ON) and Pessimistic Noise (PN)-- representing the scenarios when a human overestimates or under-estimates the cost of the other humans for performing a task. When a human overestimates the cost of others, it realizes edits to a proposed allocation will lead to a higher cost for other agents who will thus reject it. Thus, optimal counterfactual ceases to exist and thus, explanations have length zero (reducing the average length). In the case of PN, the human underestimates the cost of teammates. Thus, upon being given an allocation, they often feel this is inexplicable and find an optimal counterfactual demanding explanations. As random noise is a combination of both ON and PN (overestimates costs of some humans for particular tasks but underestimates their cost for other tasks etc.), the increase in the length of explanations is counteracted by zero length explanations. Hence, in expectation, we do not see an increase in explanation length as we increase the random noise magnitude.
As per this understanding, when we increase $\epsilon$ and only allow for PN, we clearly see an increase in the mean explanation length (shown in green).

When $\epsilon = 0$, there is no noise added to the costs, i.e. the humans have complete knowledge about the team's and the other agent's costs. Yet, due to limited computational capabilities, a human may still come up with a counterfactual that demands explanation. Hence, we observe a mean explanation length of $1$ even for zero noise. This should not be surprising because, when coming up with a foil, a human only reasons in the space of $n*(m-1)$ allocations (instead of $n^m$ allocations).
\looseness =-1
\begin{figure}[t]
\begin{tikzpicture}
    \begin{axis}[
        width=.85\linewidth,
	    height=.6\linewidth,
        major x tick style = transparent,
        ybar=2*\pgflinewidth,
        bar width=10pt,
         ymajorgrids=true,
        xmajorgrids=true,
        grid style=dashed,
        xlabel={\small $\#$ of agents about whom a human has complete knowledge}, 
        ylabel = {Rel. Exp. Length $\rightarrow$},
        symbolic x coords={$1~(33.33\%)$,$2~(66.67\%)$,$3~(100.00\%)$},
        xtick = data,
        scaled y ticks = false,
        enlarge x limits=0.25,
        ymin=0,
        legend cell align=left,
        legend style={
                at={(.96,.62)},
                anchor=south east,
                column sep=1ex
        }
    ]
        \addplot[style={Bittersweet,fill=Bittersweet!70!white,mark=none}]
            coordinates {($1~(33.33\%)$, 0.2912109375) ($2~(66.67\%)$, 0.1958984375) ($3~(100.00\%)$, 0.0966796875)};

        \addplot[style={Periwinkle,fill=Periwinkle!70!white,mark=none}]
             coordinates {($1~(33.33\%)$, 0.294921875) ($2~(66.67\%)$, 0.244921875) ($3~(100.00\%)$, 0.0966796875)};

        \addplot[style={TealBlue,fill=TealBlue!70!white,mark=none}]
             coordinates {($1~(33.33\%)$, 0.3408203125) ($2~(66.67\%)$, 0.2443359375) ($3~(100.00\%)$, 0.0966796875)};

        \legend{$\mu=1$, $\mu=3$, $\mu=5$}
    \end{axis}
\end{tikzpicture}
\vspace{-5pt}
\caption{As the number of agents about whom a human has complete knowledge (co-workers whose costs you know) increases, the mean length of explanations decreases.}
\label{fig:subset_el}
\vspace{-0.4cm}
\end{figure}

\noindent{\textit{\textbf{Incompleteness about a subset of agents.}}~~}
In many real-world scenarios, an agent may have complete knowledge about some of their team-mates but noisy assumptions about others. To study the impact of such incompleteness, we considered a modified scenario project-management domain \cite{certa2009multi} with four tasks and four humans. We then choose to vary the size of the sub-set about whom a human has complete knowledge. In Fig. \ref{fig:subset_el}, we plot the mean length of explanations, depending upon the subset size about whom the human has complete knowledge.
On the x-axis, we plot the size of the subset and on the y-axis, we show the relative explanation length that equals to explanation length divided by the longest explanation ($4^4 =256$) we can have in this setting.
We consider five runs for each sub-set size and only pessimistic noise (that ensures a high probability of having a counterfactual and thus, needing explanations).
We notice as the number of individuals about whom a human has complete knowledge increases, the mean relative explanation length (times the max explanation length) decreases uniformly across the different magnitude of noise $\mu$.
Even when a human has complete knowledge about all other agent's costs, happens whenever the size of the sub-set is $n-1$ (three in this case), it may still have some incompleteness about the team's performance costs. Added with limited computational capabilities (to search in the space of $16$ allocations), they might still be able to come up with counterfactual; in turn, needing explanations.
 Thus, the third set of bar graphs (corresponding to the label $3~(100.00\%)$ on the x-axis) has a mean of $\approx 0.1$ relative explanation length.
\vspace{-8pt}
\section{Conclusion}
\vspace{-5pt}
In this paper, we considered a task allocation problem where AITA, a centralized AI Task Allocator comes up with a negotiation-aware explicable allocation using a simulated negotiation based approach, which are popular in distributed task allocation settings, for a team of humans. When the humans have limited computational capability and incomplete information about all costs except their own, they may be dissatisfied with AITA's proposed allocation and question AITA using counterfactual allocations that they believe are explicable. We show that in such cases, AITA is able to come up with a negotiation tree that (1) is representative of the inference methodology used and (2) explains that if the counterfactual was considered, it would result in final allocation that is worse-off than the one proposed. With human studies, we show that the negotiation-aware allocation appears as fair to majority of humans while for the others, our explanations are {\em understandable} and {\em convincing}.
We also perform experiments to show that when agents either overestimate the cost of other agents or have accurate information about more agents, the average length of explanations decreases. 

\paragraph{\textbf{Acknowledgments.} } This research is supported in part by ONR grants N00014-16-1-2892, N00014-18-1- 2442, N00014-18-1-2840, N00014-9-1-2119, AFOSR grant FA9550-18-1-0067, DARPA SAIL-ON grant W911NF-19- 2-0006, and a JP Morgan AI Faculty Research grant.

\bibliographystyle{named}
\bibliography{main}
\includepdf[pages=-,fitpaper]{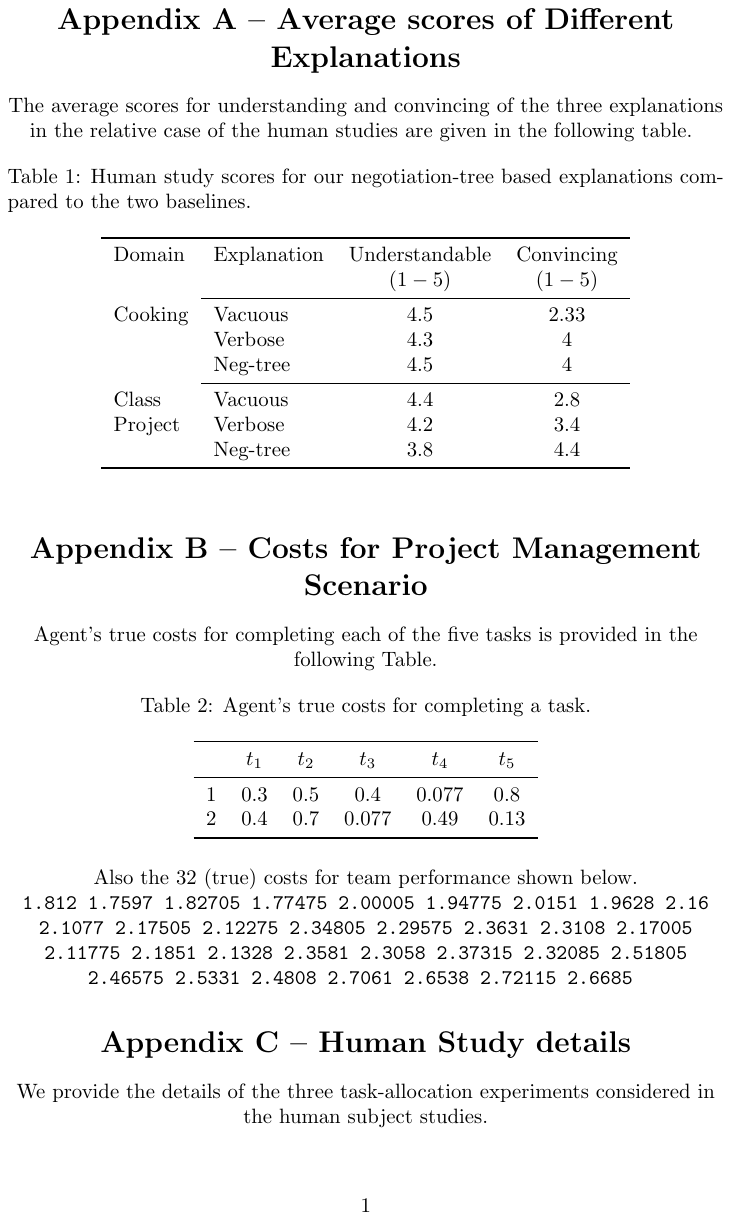}
\end{document}